\def\year{2020}\relax
\documentclass[letterpaper]{article} 
\usepackage{aaai20}  
\usepackage{times}  
\usepackage{helvet} 
\usepackage{courier}  
\usepackage[hyphens]{url}  
\usepackage{graphicx} 
\usepackage{graphicx}  
\usepackage[misc]{ifsym}
\usepackage{multirow}
\usepackage{float}
\title{Video Affective Effects Prediction with Multi-modal Fusion and Shot-Long Temporal Context}
\author{Jie Zhang\textsuperscript{\rm 1, \rm 2}, \Letter Yin Zhao\textsuperscript{\rm 1}, Longjun Cai\textsuperscript{\rm 1}, Chaoping Tu\textsuperscript{\rm 1}, Wu Wei\textsuperscript{\rm 2}\\ 
\textsuperscript{\rm 1}Alibaba Group\\
\textsuperscript{\rm 2}South China University of Technology\\ 
\{zj209798@alibaba-inc.com, auzj\_alex@mail.scut.edu.cn\}; \\
\{yinzhao.zy,longjun.clj,chaoping.tcp\}@alibaba-inc.com; weiwu@scut.edu.cn
}
 
\begin{document}
\bibliographystyle{aaai}
\maketitle

\begin{abstract}
Predicting the emotional impact of videos using machine learning is a challenging task considering the varieties of modalities, the complicated temporal contex of the video as well as the time dependency of audiences’ emotional states. Feature extraction, multi-modal fusion and temporal context fusion are crucial stages for predicting valence and arousal values in the emotional impact, but have not been successfully exploited. In this paper, we propose a comprehensive framework with novel designs of modal structure and multi-modal fusion strategy. We select the most suitable modalities for valence and arousal tasks respectively and each modal feature is extracted using the modality-specific pre-trained deep model on large generic dataset. Two-time-scale structures, one for the intra-clip and the other for the inter-clip, are proposed to capture the temporal dependency of video content and emotion states. To combine the complementary information from multiple modalities, an effective and efficient residual-based progressive training strategy is proposed. Each modality is step-wisely combined into the multi-modal model, responsible for completing the missing parts of features. With all those improvements above, our proposed prediction framework achieves better performance on the LIRIS-ACCEDE dataset with a large margin compared to the state-of-the-art.
\end{abstract}

\section{Introduction}
\noindent Affective video content analysis aims at predicting the videos' emotional impact on audiences. It plays important roles in understanding the videos' content, highlight detection, and fundamental support for several advanced applications such as multi-media search with sentimental query world. Furthermore, the rapid growth of the movie industry brings a large number of movies. Evaluating the movies' quality and online-effect is essential for online media-service providers such as Youku(Alibaba), Netflix, etc.. Predicting the audiences' emotional evolvement when watching movies is an important way to help both online media-server providers or filmmakers to invest movies, evaluate on-line effect as well as distribute them more efficiently, etc.

In the affective computing community, human emotions can be categorically or continuously defined \cite{izard2007basic,barrett2007mice}. Emotion categories are happy, sad, angry, surprise, disgust, neutral, which are commonly used in emotion recognition or classification \cite{cowie2001emotion}. Compared to the categorical definition, dimensional definition describes the emotions continuously in two dimensions: Valence (positive vs. negative) and Arousal (active vs. calm). Any human emotion can be located in the space spanned by the two dimensions, which is more fine-grained than the categorical definition. The goal of our task is to predict the audiences' emotional states based on the movie content, i.e. the valence and arousal values with the movie going on. 

Finding discriminative features from raw videos for predicting valence and arousal values is far away from an easy task. Video is typical multi-modal media, involving audio and visual content. Even if in visual content, human facial expressions, pose behaviors, scenes, etc. can also be regarded as modalities. Audiences' emotions can be triggered by any modality such as the actors' expressions or actions, the movies' scenes (environment, atmosphere) as well as background music. Therefore, the mainstream of affective video content analysis is to extract multi-modal features and combine all those features. In early stages, most works \cite{xu2013hierarchical,moreira2015recod} extract handcrafted features such as Local Binary Patterns (LBP), Histogram of oriented gradient (HOG), etc., for representing visual signals, and Linear Predictive Coding coefficients (LPC), Mel Frequency Cepstral Coefficient (MFCC) for representing audio signals. Recently, deep models are introduced \cite{liu2018multi} to retrieve high-level features with semantic meanings, bringing more chances to exploit the problems. To better capture the videos' affective content, we systematically analyze the importance of every modality and select the most suitable modalities for each task. Then, deep models pre-trained on larger datasets are used to extract multi-modal features. 

Feature fusion is another challenging step. Multi-modal features are always complementary and the importance of each modality dynamically changes over time. For example, some movie clips' emotional impact can be captured by audio content while others may rely on visual features. Current studies of affective video content analysis mainly adopt either decision-level fusion \cite{dobrivsek2013towards} or feature-level fusion \cite{wimmer2008low}. The former combines results from each modality through voting or weighted average methods. Each modality-specific model is trained independently which can't exploit the fusion of multi-modal features. The latter concatenates multi-modal features and learn parameters for all modalities at the same time which can easily lead to overfitting. We design a progressive training algorithm where each modality is trained and fine-tuned stepwisely. Each modality is only responsible for completing the missing parts of features extracted from previous modalities, thus the most discriminative modalities can be dynamically selected for each movie clip and the complementarity of multi-modal features can be fully utilized. The overfitting risk can also be suppressed since fewer parameters are learned at each step.

Besides all the above, we also investigate how to utilize the temporal context of videos for sentiment prediction, which is lacking in most of the existing works in affective video content analysis, where they simply apply LSTMs \cite{hochreiter1997long} or GRUs \cite{cho2014learning} for temporal dependency. We propose two time-scale model structures considering the video's shot-long temporal context. For the shot-time context, LSTMS are used for each modality. For the long-time dependency of valence task, a structure that is similar to the temporal segment network (TSN) \cite{wang2016temporal} is used to capture the long temporal context. For the arousal task, a moving mean post-processing method is adopted to utilize the trend of previous emotions.

The contributions of this paper are as follows:
\begin{itemize}
\item We propose an effective feature extracting framework. The most suitable modalities are selected for each task and deep models trained on each modality with much larger datasets are used for extracting multi-modal features to represent the emotional video content.
\item We design two-time-scale model structure considering the video's shot-long temporal context. A TSN-like structure and a moving average post-processing method are proposed to effectively fuse the temporal context for valence and arousal tasks respectively.
\item We design a residual-based progressive training algorithm for multi-modal fusion. A multi-modal model is trained through combining each modality step-wisely and each modality is only responsible for completing the missing parts of features extracted from previous modalities. Thus it can choose the most discriminative modalities for each movie clip.
\end{itemize}

\section{Related Works}
Video content involves both audio and visual elements. The basic part of affective video content analysis involves extracting audio and visual features to characterize the video content. In early stage, hand-crafted features play important roles in feature extracting. In recent years, with the development of deep learning, semantics-meaningful features extracted by deep models are becoming more and more popular in affective video content analysis and many works extract features with pre-trained CNN models. For example, A VGG-like model named VGGish \cite{hershey2017cnn} is adopted for extracting audio features and CNN models trained on generic task datasets, such as ImageNet \cite{deng2009imagenet}, RAF \cite{li2017reliable} are used for extracting visual features \cite{liu2018multi}. 

Another focus of the emotion prediction is multi-modal fusion \cite{atrey2010multimodal}. Multi-modal fusion methods can be divided into two categories: feature-level fusion, and decision-level fusion. The key difference between the above two categories is the stage when fusion happens. Rosas, Mihalcea, and Morency \shortcite{rosas2013multimodal} concatenate linguistic, audio and visual features into a common feature vector and seek to find the hyperplane that best separates positive examples from negative examples using SVMs \cite{cortes1995support} with linear kernels. Wang and Cheong \shortcite{wang2006affective} characterized every scene through concatenating audio and visual features to a vector and adopt a specially adapted variant of SVM to recognize anger, sadness, surprise, happiness, disgust and neutral. Metallinou, Lee, and Narayanan \shortcite{metallinou2010decision} model face, voice and head movement cues for emotion recognition and fuse the results of all classifiers using a Bayesian framework. While those models can outperform unimodal models, they fail to utilize the dependencies among different modalities. Pang, Zhu, and Ngo \shortcite{pang2015deep} use Deep Boltzmann Machine (DBM) to learn the highly non-linear relationships that exist among low-level features across different modalities for emotion prediction. Gan et al. \shortcite{gan2017multimodal} propose a multi-modal deep regression Bayesian network (MMDRBN) to capture the dependencies between visual elements and audio elements and a fast learning algorithm is designed to learn the regression Bayesian network (RBN), Then the MMDRBN is transformed into an inference network by minimizing the KL-divergence.

To utilize the temporal context, Kurpukdee et al. \shortcite{kurpukdee2017speech} extract phoneme-based features from raw input speech signals using convolutional long short-term memory (LSTM), recurrent neural network (ConvLSTM-RNN) and adopt support vector machines (SVM) or linear discriminant analysis (LDA) to classify four emotions (anger, happiness, sadness and neutral). Fan et al. \shortcite{fan2016video} use RNNs to fuse features extracted by the convolutional neutral network (CNN) over individual video frames and use C3D \cite{tran2015learning} to encode appearance and motion information at the same time, then the predicted scores from different models are combined in a weighted-sum rule.

Those previous studies only focus on a part of the three aspects mentioned above, while all aspects are necessary to get a good performance. Different from those works, we make a deep analysis of all three aspects and propose effective methods for each of them. With all three aspects considered, our emotion prediction framework gets better performance compared with other works.

\section{Proposed Methods}
\begin{figure*}[htb]
	\centering
	\includegraphics[width=0.9\textwidth]{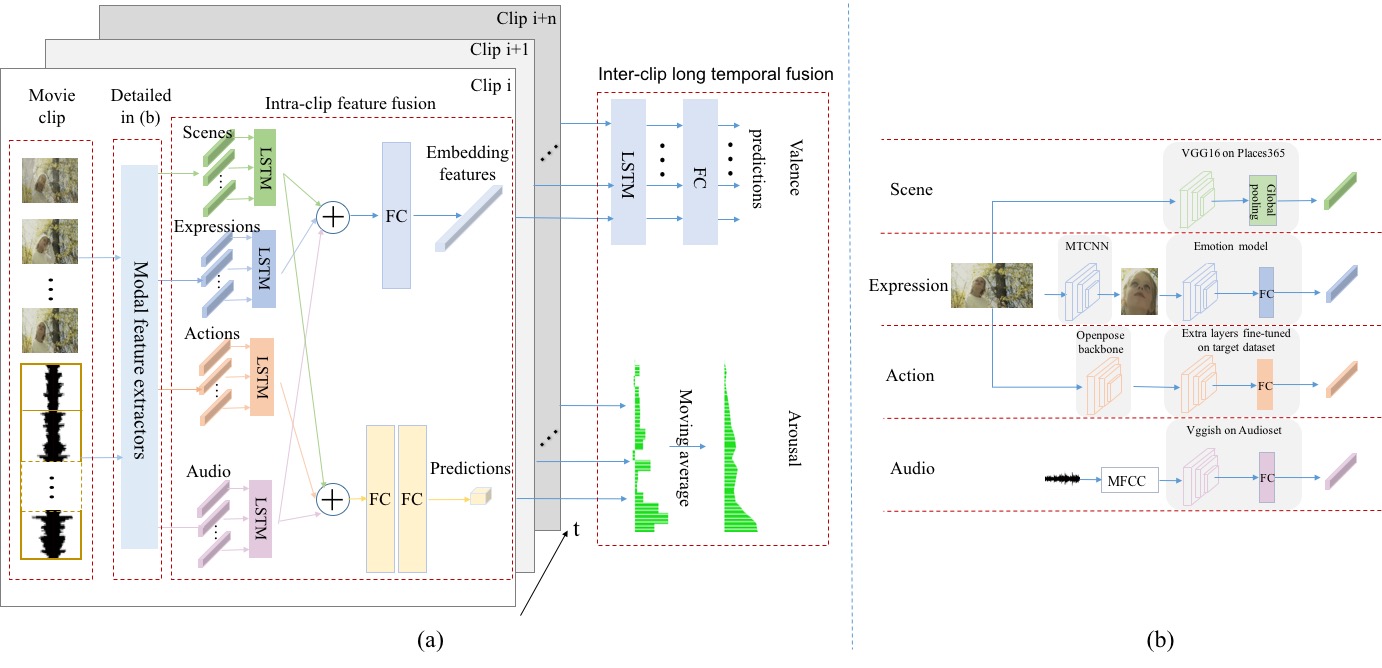} 
	\caption{Part (a) describes our overall framework for valence and arousal tasks separately. Our framework involves two parts: intra-clip part and inter-clip part. First, we divide movies to continuous clips. For each clip, features for multiple modalities are extracted as detailed in part (b). Then LSTMs are used to fuse short temporal information and feature-level fusion for multi-modal features is performed. Two different methods are used to combine context information among continuous clips for valence and arousal respectively.}
	\label{fig1}
\end{figure*}

We divide the untrimmed movies to non-overlap short clips with same length and predict valence and arousal for each clip. The overall framework consists of three major steps as shown in Fig. \ref{fig1}. First, we extract frame-level modality-specific features using commonly-used pre-trained audio and image feature extractors. Then, the intra-clip feature fusion and inter-clip context fusion are used to consider the short and long temporal context. For intra-clip feature fusion, we append LSTMs after the feature extractor for each modality to consider the short-time temporal relation within the clip. Then, all modality-specific features are summed into a vector, representing the clip-level multi-modal features. For the inter-clip context, we adopt LSTMs based on the clip-level features to capture the long-time dependency between clips for the valence task. For the arousal task, a more simple but effective exponential moving average with decay weights is utilized. A residual-based progressive training algorithm is designed to train the network.

\subsection{Multi-modal Features}
The modal scope we could use for valence and arousal tasks in movie affective analysis is based on the following observations. 1) The actors/Actress' actions, dialogues, and facial expressions are the key factors that affect the audience's emotions. Scenes (such as environment, atmosphere) and background music also implicitly deliver emotions to audiences. For example, audiences can feel the role's anger through his loud voice and feel peace when listening to the music with a soothing rhythm. 2) Valence and Arousal emotional states depend on different modalities. For instance, Facial expressions are more related to the valence task while actions are more related to the arousal task. Audiences generally feel sad when actors cry and feel happy when actors laugh, but the emotional intensity does not differ much. However, people often feel excited when actors in the video behave dramatically, such as struggling or fighting and feel calm when actors behave smoothly no matter which direction the valence is. This demonstrates that the actors' actions can affect the intensity of emotions while they are less related to the direction of emotions \cite{detenber1996bio}. Therefore, for the valence task, audio, scene and facial expression modalities are finally used. For the arousal task, audio, scene, and action modalities are finally used.

\subsubsection{Audio Features}
Audio information in a movie can be divided into two main parts: the physical characteristics of sound and the content contained by sound. The physical characteristics of sound consist of intensity, speed, rhythm, etc. while the audio content involves complex language understanding. We only focus on language-independent audio features. Here we adopt VGGish \cite{hershey2017cnn} to extract semantically meaningful features with all audio characteristics taken into account itself. The inputs are log Mel-spectrograms with shape 96x64 computed with 0.96s long audio clips. We adopt the first 960 ms from every second of the audio. We adopt the output of the 128-wide fully connected layer followed by a PCA transformation and quantization as the compact embedding features for audio. Thus, every second of audio is converted into a 128-dimension vector. 

\subsubsection{Visual Features}
Visual features consist of the human' actions, expressions as well as scenes in videos. First, a sparse sampling strategy, one frame per second, is adopted to sample frames from clips to reduce noise introduced by highly similar frames and preserves relevant information with a dramatically lower cost. Then each feature will be extracted in frame-level. 

We adopt pose-related features to represent the human' actions. Specifically, we append two groups of convolution/maxpool layers and a 128-wide fully connected layer after the last convolution layer of the pre-trained OpenPose's backbone \cite{cao2018openpose}. The weights of the OpenPose's backbone are fixed and the newly-added layers will be fine-tuned on the target dataset. 
To represent the actors' expressions, we pre-train an Xception network \cite{chollet2017xception} with fully connected layers on RAF to extract emotional facial features. To extract facial expression feature, we first detect faces from frames by MTCNN \cite{zhang2016joint}. Then we crop the largest face detected and resize it to 160x160. The face image is fed into the pre-trained model and is represented by a 3072-dimension vector extracted from the last fully connected layer. If no faces are detected, the average face across the whole training dataset is used. 
A pre-trained VGG16 network on Places365 is adopted to extract features of the movies' scenes. We resize the frames to 224x224 and extract 512-dimension features at the last pooling layer. 

\subsection{Feature Fusion With Shot-long Temporal Context}
\subsubsection{Intra-Clip Short Temporal Fusion}
Since each modality has its own time-dependency, we first carry out short temporal fusion for each modality.
Based on the previous section, frame-level features for each modality can be obtained. Then we adopt two layers of bidirectional LSTMs for each modality inside the clip to fuse the temporal information. The hidden state of the final step of the LSTMs is adopted as the clip-level modality-specific features.
Having obtained the modality-specific features, we first sum all the features together for each clip. Then a fully connected layer is followed, forming the final clip-level multi-modal features.

\subsection{Inter-clip Long Temporal Fusion}
We adopt two different methods to utilize the temporal context among clips for valence and arousal tasks. 

\subsubsection{Valence}
We adopt a TSN-like structure to utilize the temporal context among continuous clips. Specifically, we use two-layers bidirectional LSTMs to combine the temporal context among continuous clips. The input for each step is the clip-level multi-modal features obtained in the intra-clip short temporal fusion.
The hidden state for each step is the final embedding features combining long temporal context for the corresponding clip. A fully connected layer with tanh activation is adopted to make a final valence value prediction for each clip.

\subsubsection{Arousal}
For arousal value prediction, we directly append another fully connected layer with tanh activation as the raw arousal prediction after the intra-clip features.
Then a simple but effective exponential moving average with decay weights is utilized to consider the temporal context. The reason we use moving average instead of LSTMs like the valence task is that the arousal value represents the intensity of emotions, which cannot dramatically change in relative short time, while for valence, the emotion direction might be more context-related.

\subsection{Training Strategy}
Training the whole network, which has multiple stages with several sub-structures, in an end-to-end way is not an easy task, which can easily lead to be over-fitting because of the large number parameters. Thus, we train the whole network part by part, i.e. first for the intra-clip feature fusion part and then inter-clip long temporal fusion part. For VGGish, facial expression model, VGG16 pre-trained on Places365, and modified OpenPose model, the weights are always fixed in our tasks.

\subsubsection{Residual-based Progressive Training Strategy}
\begin{figure*}[htb]
	\centering
	\includegraphics[width=0.9\textwidth]{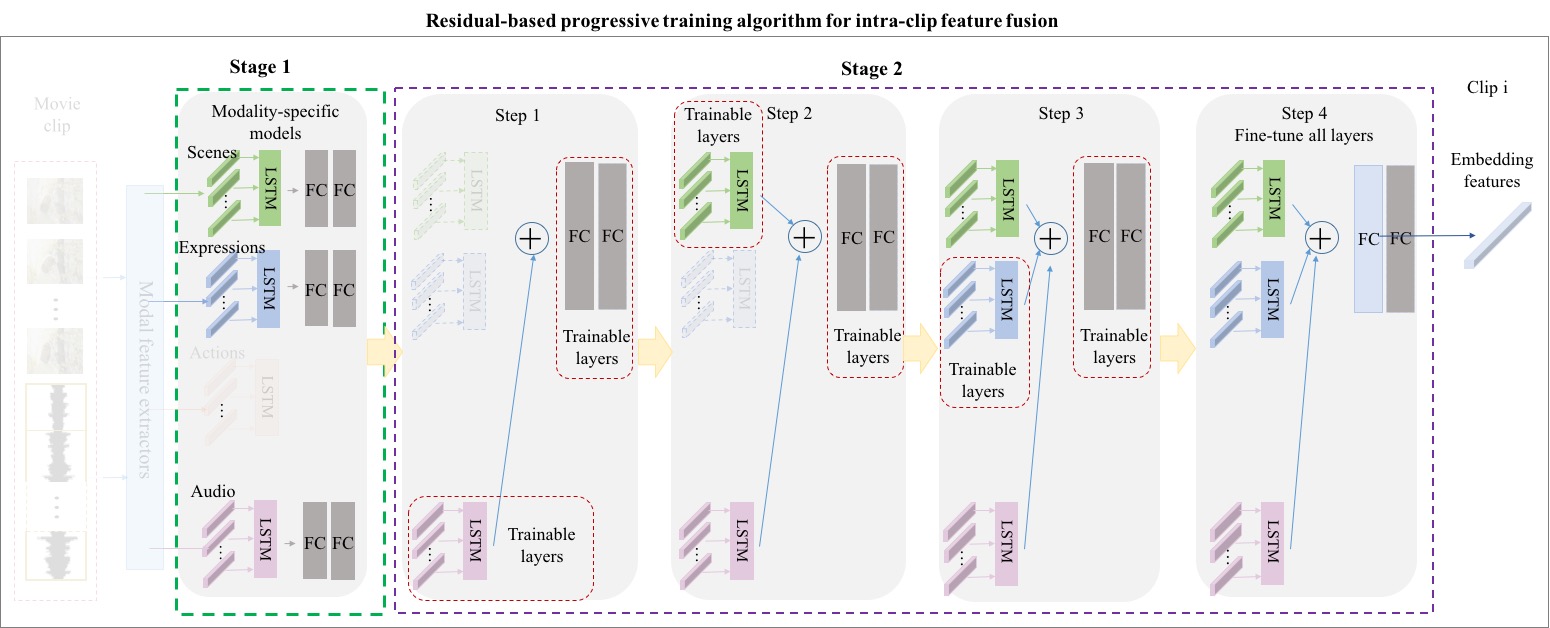} 
	\caption{Residual-based progressive training strategy for intra-clip feature fusion of valence}
	\label{fig2}
\end{figure*}

To learn the LSTMs for every modality in short temporal fusion and effectively fuse multi-modal features, we design a two-stage residual-based progressive training strategy as shown in Fig \ref{fig2}.

\noindent \textit{Stage 1: Modality-specific pre-training}

The main role of this step is to obtain the importance of each modality and determine the training order for each modality in the next stage. We append two fully connected layers for auxiliary training after the LSTMs of each modality to form the modality-specific model. Then we train each modality-specific model by minimizing the Mean Square Error (MSE) loss. The loss function can be computed as follows:
\begin{equation}
L(y,G)=\sum \limits_{j=1} ^{m}\frac{1}{m}||y_{j}-G_{j}||_{2} ^{2} \label{eq1}
\end{equation}

Where $y_{j}$ is the prediction for clip $j$. $G_{j}$ represents the ground truth for clip $j$. It is the mean of labels for every second within the clip. $m$ denotes the batch size.

After the models are trained, we sort modalities in descending order according to their performance. Thus, modality $i$ means the modality with the $ith$ high performance and the LSTMs for it will be trained in the sequence of $i$. 

\noindent \textit{Stage 2: Training intra-clip model}

At the first step, we train the LSTMs for modality 1. At the second step, we sum the features extracted by the LSTMs for modality 1 and modality 2 and only train the LSTMs for modality 2. The remaining modalities' LSTMs are learned similarly. At each step, two fully connected layers are appended after the summed features to make predictions. The fully connected layers are used for auxiliary training and only exist during the training phase. After we get the weights for all modalities' LSTMs, we add two fully connected layers after the LSTMs and fine-tune the entire model. For the valence task, the output of the first fully connected layer is used as the clip-level embedding features. For the arousal task, the fully connected layers are used to make the prediction for each clip. The training process for the valence task is shown in Fig. \ref{fig2} as an example.

With the residual-based progressive training strategy described above, the model can dynamically select important modalities. Formally, assuming that we are training the $ith$ modality, the features combining the first $i-1$ modalities are denoted as $f_{i-1}$. The desired features that can get the best performance with the $ith$ modality combined are denoted as $f_{i}$. Thus the LSTMs for modality $i$ fit the mapping $H_{i}(x):=f_{i}-f_{i-1}$. If $f_{i-1}$ is discriminative enough to make correct predictions and modality $i$ is of low importance, it would push the mapping to zero. If modality $i$ plays an important role, it would push the mapping to complete the $f_{i-1}$ towards $f_{i}$ to get better performance.

\subsubsection{Long Temporal Fusion}
For arousal, there is no parameters to train for long temporal fusion since we adopt an exponential moving average post-processing method and predictions are made in the intra-clip part. Formally, the final prediction can be computed as follows:
\begin{equation}
ema_{i}=ema_{i-1}*\beta + y_{i}*(1-\beta) \label{eq3}
\end{equation}

Where $y_{i}$ is the prediction for clip $i$. $ema_{i}$ denotes the exponential moving average value for clip $i$ which is used as the final prediction. $\beta$ is the weight decay.

For valence, we keep the weights for the intra-clip part fixed and only train the LSTMs and fully connected layer for the inter-clip part. The loss function can be computed as follows:
\begin{equation}
L(Y,G)=\sum \limits_{j=1} ^{m}\frac{1}{m*L}\sum \limits_{i=1} ^{L}||y_{j}^{i}-G_{j}^{i}||_{2} ^{2} \label{eq2}
\end{equation}

Where $y_{j}^{i}$ and $G_{j}$ denote the prediction and ground truth for the $ith$ clip in the $jth$ example respectively. $m$ represents the batch size. $L$ is the number of clips in an example.

\section{Experiments and Results}
\subsection{Dataset and Metrics}
The LIRIS-ACCEDE dataset is the largest dataset for affective video content analysis, which is used in the MediaEval 2018 emotional impact of movies task. The LIRIS-ACCEDE dataset contains videos from a set of 160 professionally made and amateur movies. Several movie genres are represented in this collection of movies such as horror, comedy, drama, action and so on. Languages are mainly English with a small set of Italian, Spanish, French, and others. A total of 54 movies (total duration of 26 hours and 49 minutes) from the set of 160 movies are provided as the development set. 12 other movies (total duration of 8 hours and 56 minutes) consist the test set. The scores of valence and arousal which range from -1 to 1 are provided continuously (every second) along movies. Valence is defined on a continuous scale from most negative to most positive emotions, while arousal is defined continuously from calmest to most active emotions.

The official metric is the Mean Square Error (MSE), which is the common measure generally used to evaluate regression models. However, we also consider Pearson's Correlation Coefficient (PCC) for the emotional trend analysis of movies.

\subsection{Results}
In our experiments, movies are converted into continuous non-overlap clips with duration 10s. In the test phase, we obtain the entire movie's predictions by concatenating the predictions for all clips. Since the predictions are clip-level, to get predictions for every second, we repeat every clip-level prediction 10 times.

\subsubsection{Modality-specific Performance}
To evaluate the importance of every modality for each task, we train modality-specific models. The performance of each model is shown in Table. \ref{tab1}. Besides, to better demonstrate each modality's effectiveness, we visualize the features extracted from clips in the test set by the modality-specific models. T-SNE is adopted for dimensionality reduction.

\begin{table}[htb]
	\caption{Performance for modality-specific models.}\smallskip
	\centering
	\resizebox{.95\columnwidth}{!}{
		\smallskip\begin{tabular}{c|c|c|c|c}
			\hline
			\multirow{2}{*}{Modality used} &\multicolumn{2}{c|}{Valence}&\multicolumn{2}{c}{Arousal}\\
			\cline{2-5}
			{}&MSE&PCC&MSE&PCC\\
			\hline
			Audio&0.098&0.264&0.140&0.172\\
			Scene&0.103&0.192&0.152&0.140\\
			Expression&0.110&0.150&0.162&0.061\\
			Action&0.132&0.057&0.156&0.158\\
			\hline
		\end{tabular}
	}
	\label{tab1}
\end{table}

\begin{figure}[htb]
	\centering
	\includegraphics[width=0.9\columnwidth]{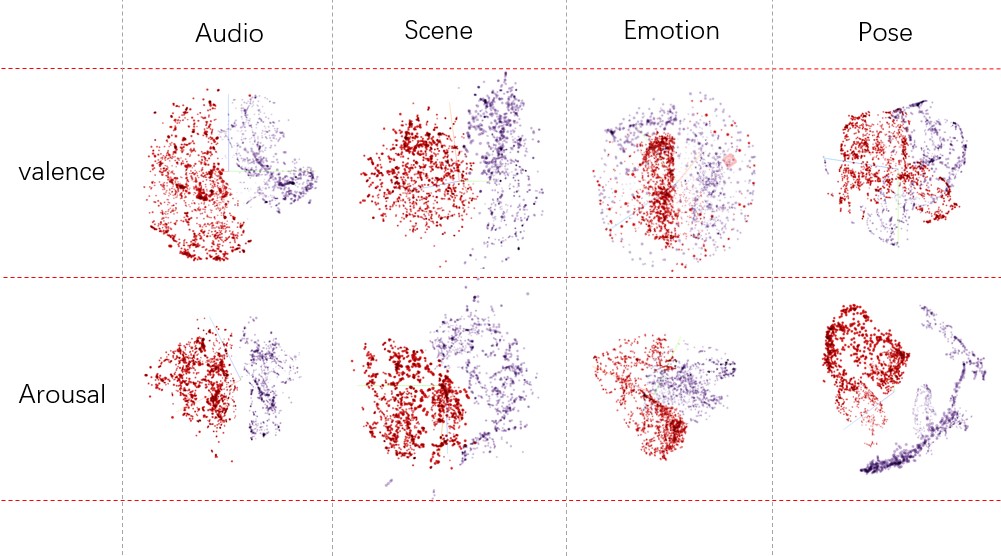} 
	\caption{Features extracted from movie clips in the test set by the modality-specific models. T-SNE is adopted for dimensionality reduction. The red points represent the features with label greater than 0 and the blue points represent the features with label less than 0.}
	\label{fig3}
\end{figure}

As shown in Table. \ref{tab1}, modality-specific model trained on audio features has the best performance for both valence and arousal tasks. This demonstrates that audio signals contain more emotional information. For the valence task, expressions and scenes play similar importance while actions have the worst performance. For the arousal task, actions and scenes have equal status while expressions are useless. The importance of each modality can also be revealed by the discriminative of features extracted through the modality-specific models as shown in Fig. \ref{fig3}. This demonstrates our analysis in section "Multi-modal Features".

\subsubsection{Performance for Intra-clip Part}
\begin{figure}[htb]
	\centering
	\includegraphics[width=0.6\columnwidth]{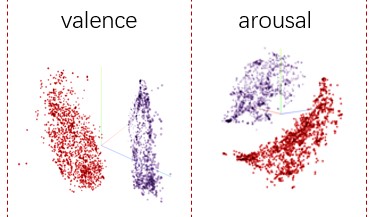} 
	\caption{Features extracted from movie clips in the test set by the multi-modal models. T-SNE is adopted for dimensionality reduction. The red points represent the features with label greater than 0 and the blue points represent the features with label less than 0.}
	\label{fig4}
\end{figure}

According to the analysis and experiments, audio and scenes are used for both valence and arousal tasks while expressions are only used in the valence task and actions are only used in the arousal task. The training process is described in section "Residual-based Progressive Training Strategy". We also conduct two comparative experiments: feature-level fusion with traditional training method, training the entire model at the same time, which is referred as M1, and decision-level fusion, learn modality-specific models independently and average the results, which is referred as M2. For the valence task, M1 and M2 use audio, scenes, and expressions. For the arousal task, M1 and M2 use audio, scenes, and actions. The results of various methods are listed in Table. \ref{tab2}. Besides, we also visualize the features obtained by the multi-modal fusion through T-SNE, which is shown in Fig. \ref{fig4}.

\begin{table}[htb]
	\caption{Performance for multi-modal models.}\smallskip
	\centering
	\resizebox{.95\columnwidth}{!}{
		\smallskip\begin{tabular}{c|c|c|c|c}
			\hline
			\multirow{2}{*}{} &\multicolumn{2}{c|}{Valence}&\multicolumn{2}{c}{Arousal}\\
			\cline{2-5}
			{}&MSE&PCC&MSE&PCC\\
			\hline
			M1&0.104&0.284&0.148&0.257\\
			M2&0.090&0.300&0.142&0.278\\
			Audio+Scene&0.091&0.348&-&-\\
			Audio+Action&-&-&0.140&0.293\\
			Audio+Scene+Expression&0.089&0.358&-&-\\
			Audio+Action+Scene&-&-&0.138&0.314\\
			\hline
		\end{tabular}
	}
	\label{tab2}
\end{table}

As shown in Table. \ref{tab2}, feature-level fusion with our residual-based progressive training algorithm get better performance compared with traditional training method and also outperforms the decision-level fusion. Fig. \ref{fig4} shows that the features extracted by the multi-modal models are discriminative and emotion-related. This demonstrates the effectiveness of our training algorithm to fully use the complementary of multi-modal features.

\subsubsection{Performance for Long Temporal Fusion}
The methods to utilize the long temporal context are described in section "Long Temporal Fusion". We try different clips number for the valence task and different decay weights $\beta$ for the arousal task to get the best performance.

\begin{table}[htb]
	\tiny
	\caption{Performance for the valence task with long time information.}\smallskip
	\centering
	\resizebox{.7\columnwidth}{!}{
		\smallskip\begin{tabular}{c|c|c}
			\hline
			number of clips&MSE&PCC\\
			\hline
			3&0.073&0.406\\
			4&0.071&0.444\\
			5&0.072&0.419\\
			6&0.077&0.380\\
			\hline
		\end{tabular}
	}
	\label{tab3}
\end{table}
\begin{table}[H]
	\tiny
	\caption{Performance for the arousal task with long time information.}\smallskip
	\centering
	\resizebox{.7\columnwidth}{!}{
		\smallskip\begin{tabular}{c|c|c}
			\hline
			decay weights $\beta$&MSE&PCC\\
			\hline
			0.96&0.136&0.400\\
			0.97&0.136&0.409\\
			0.98&0.137&0.419\\
			0.99&0.140&0.427\\
			\hline
		\end{tabular}
	}
	\label{tab4}
\end{table}

As shown in Table. \ref{tab3} and Table. \ref{tab4}, the proposed methods to use long temporal information for valence and arousal tasks can significantly promote performance. For the valence task, using 4 continuous movie clips gets the best result. Using fewer clips can't get enough long time information while using more clips weakens the flow of information and may introduce noises by obvious emotional change during clips. For the arousal task, experiment results show that the moving mean post-processing method can significantly increase PCC by modifying previous emotional trends according to the current prediction. The increase in MSE with larger momentum is because the arousal value of every second is pulled towards the mean and may be further from the ground truth. 

\begin{table}[htb]
	\caption{Performance compared with the state-of-the-art.}\smallskip
	\centering
	\resizebox{.95\columnwidth}{!}{
		\smallskip\begin{tabular}{c|c|c|c|c}
			\hline
			\multirow{2}{*}{} &\multicolumn{2}{c|}{Valence}&\multicolumn{2}{c}{Arousal}\\
			\cline{2-5}
			{}&MSE&PCC&MSE&PCC\\
			\hline
			CERTH-ITI \cite{CERTH-ITI}&0.117&0.098&0.138&0.054\\
			THUHCSI \cite{jin2018thuhcsi}&0.092&0.305&0.140&0.087\\
			Quan, Nguyen, and Tran \shortcite{QUAN}&0.115&0.146&0.171&0.091\\
			Yi, Wang, and Li \shortcite{Yi}&0.090&0.301&0.136&0.175\\
			GLA \cite{Sun}&0.084&0.278&\textbf{0.133}&0.351\\
			Ko et al, \shortcite{Ko}&0.102&0.114&0.149&0.083\\
			Ours&\textbf{0.071}&\textbf{0.444}&0.137&\textbf{0.419}\\
			\hline
		\end{tabular}
	}
	\label{tab5}
\end{table}

\subsubsection{Comparison with The State-of-the-art}
As shown in Table 5, our whole emotion prediction pipelines get significantly better performance compared with the other works.

Besides, we also adopt experiments on classification task of the LIRIS-ACCEDE dataset which is used in MediaEval 2015 \cite{sjoberg2015mediaeval} with the intra-clip part. The better performance compared with other works as shown in Table 6 further demonstrates the effectiveness of our extracted features and residual-based progressive training algorithm.

\begin{table}[htb]
	\tiny
	\caption{Performance compared with MediaEval 2015 related works.}\smallskip
	\centering
	\resizebox{.9\columnwidth}{!}{
		\smallskip\begin{tabular}{c|c|c}
			\hline
			{} $\beta$&Valence (acc)&Arousal (acc)\\
			\hline
			MIC-TJU \cite{mictju}&0.420&0.559\\
			NII-UIT \cite{vu2015shin}&0.430&0.559\\
			ICL-TUM-PASSAU \cite{trigeorgis2015icl}&0.415&0.557\\
			Fudan-Huawei \cite{dai2015fudan}&0.418&0.488\\
			TCS-ILAB \cite{chakraborty2015tcs}&0.357&0.490\\
			UMons \cite{seddati2015umons}&0.373&0.524\\
			RFA \cite{mironica2015rfa}&0.330&0.450\\
			KIT \cite{marin2015kit}&0.385&0.519\\
			Ours&\textbf{0.459}&\textbf{0.575}\\
			\hline
		\end{tabular}
	}
	\label{tab6}
\end{table}

\begin{figure}[H]
	\centering
	\includegraphics[width=0.9\columnwidth]{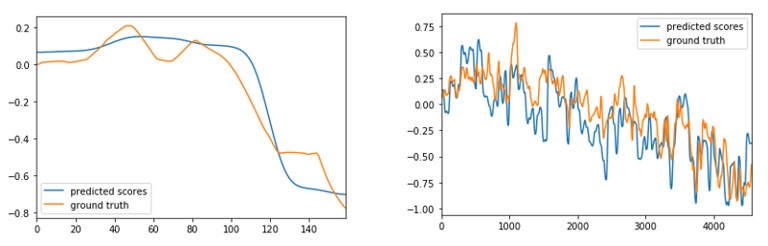} 
	\caption{The predictions of movies in test set for the valence task. The left one is for movie MEDIAEVAL18\underline{\hspace{0.5em}}54; the right one is for movie MEDIAEVAL18\underline{\hspace{0.5em}}62.}
	\label{fig5}
\end{figure}

\subsubsection{Cases Analysis}

\begin{figure}[htb]
	\centering
	\includegraphics[width=0.9\columnwidth]{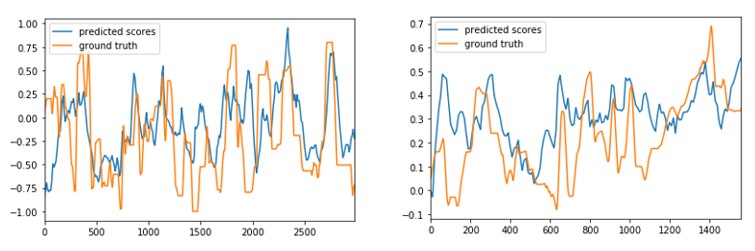} 
	\caption{The predictions of movies in the test set for the arousal task. The left one is for movie MEDIAEVAL18\underline{\hspace{0.5em}}60; The right one is for movie MEDIAEVAL18\underline{\hspace{0.5em}}63.}
	\label{fig6}
\end{figure}
\begin{figure}[htb]
	\centering
	\includegraphics[width=0.95\columnwidth]{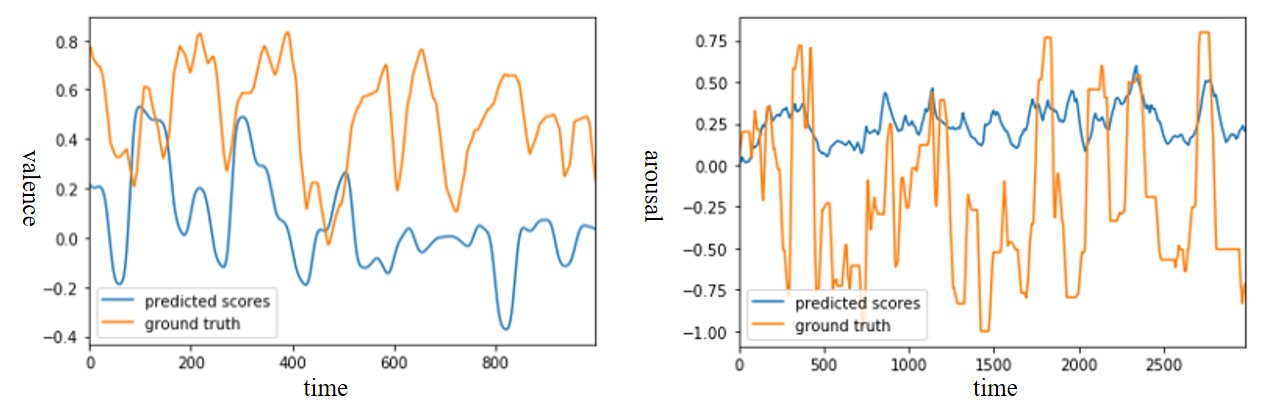} 
	\caption{Bad cases in the test set. The left one is the valence predictions for movie MEDIAEVAL18\underline{\hspace{0.5em}}65 from 3000s to 4000s; The right one is the arousal predictions for movie MEDIAEVAL18\underline{\hspace{0.5em}}60.}
	\label{fig7}
\end{figure}

As shown in Fig. \ref{fig5} and Fig. \ref{fig6}, our emotion prediction pipelines can precisely predict the trend of movies' emotional impact for valence and arousal tasks. There is some wrong prediction inevitably. To locate the problems more precisely and provide directions for future works, we further analyze the bad cases.

In the left chart of Fig. \ref{fig7}, the ground truth for many parts are positive while the corresponding predictions are negative. According to our observation, this movie is a comedy. In some movie clips, the fighting and quarrels have positive emotional impact on audiences while they are difficult to recognize for models. In the right chart, the fluctuation between the predictions and ground truth is consistent and the movie clips with high arousal labels can be accurately detected, while the movie clips with low arousal labels are predicted around 0. The reason is that for movie clips with high arousal labels, there are often fighting or loud sound which are easy to recognize, while the movie clips with low or neutral arousal labels all have similar peaceful features, their correct predictions depend on the emotional thinking way of human, which is difficult to learn. This is also the main reason why the MSE for the arousal task is much higher than that for the valence task.

According to the above analysis, there are two main directions to get better performance: Effectively use longer time information even the whole movie's contexts and better model the emotional thinking way of the human. 

\section{Conclusion and Future Work}
In this work, we propose a complete emotion prediction framework. For valence and arousal tasks, modalities are carefully selected through evaluating the importance of every modality to reduce noise introduced by less task-related modalities, then pre-trained models are used to extract semantically meaningful features. To combine multi-modal features, an effective and efficient residual-based progressive training algorithm is proposed. Besides, Two-time-scale structures valence and arousal tasks are adopted to use the shot-long temporal context. The experimental results on LIRIS-ACCEDE dataset with several comparative studies demonstrate the effectiveness of our methods. Future work might involve how to incorporate more sophisticated time-dependency modals of human emotion states and explore more discriminative features etc..

\bigskip
\bibliography{reference}
\end{document}